\pdfoutput=1

\documentclass[11pt]{article}

\usepackage{authblk}
\usepackage[]{acl}

\usepackage{times}
\usepackage{latexsym}

\usepackage{graphicx}
\usepackage{booktabs}

\usepackage[T1]{fontenc}

\usepackage[utf8]{inputenc}

\usepackage{microtype}

\newcommand{\correct}[1]{}

\title{PPL-MCTS: Constrained Textual Generation Through Discriminator-Guided MCTS Decoding}

\author[1,2]{Antoine Chaffin}
\author[1]{Vincent Claveau}
\author[1]{Ewa Kijak}
\affil[1]{CNRS, IRISA, Univ. Rennes 1, Campus de Beaulieu, 35000 Rennes, France}
\affil[2]{IMATAG, 13 Rue Dupont-des-Loges, 35000 Rennes, France}
\affil[ ]{\texttt {\{antoine.chaffin, vincent.claveau, ewa.kijak\}@irisa.fr}}

\begin{document}
\maketitle
\begin{abstract}
Large language models (LM) based on Transformers allow to generate plausible long texts. 
In this paper, we explore how this generation can be further controlled at decoding time to satisfy certain constraints (e.g. being non-toxic, conveying certain emotions, using a specific writing style, etc.) without fine-tuning the LM.
Precisely, we formalize constrained generation as a tree exploration process guided by a discriminator that indicates how well the associated sequence respects the constraint. 
This approach, in addition to being easier and cheaper to train than fine-tuning the LM, allows to apply the constraint more finely and dynamically.
We propose several original methods to search this generation tree, notably the Monte Carlo Tree Search (MCTS) which provides theoretical guarantees on the search efficiency, but also simpler methods based on re-ranking a pool of diverse sequences using the discriminator scores. 
These methods are evaluated, with automatic and human-based metrics, on two types of constraints and languages: review polarity and emotion control in French and English. We show that discriminator-guided MCTS decoding achieves state-of-the-art results without having to tune the language model, in both tasks and languages. 
We also demonstrate that other proposed decoding methods based on re-ranking can be really effective when diversity among the generated propositions is encouraged.
\end{abstract}

\section{Introduction}

Generative language models exist for a long time, but with advent of the transformer architecture~\citep{DBLP:conf/nips/VaswaniSPUJGKP17} and increasing computing capabilities, they are now able to generate well written and long texts. 
In particular, large models, such as the well known GPT-2~\citep{Radford2019LanguageMA} and GPT-3 \citep{DBLP:conf/nips/BrownMRSKDNSSAA20}, have been used successfully for various applications: assisting writers, summarizing, augmentating data for subsequent NLP tasks, generating fake news  \citep{DBLP:journals/corr/abs-2003-02245,DBLP:journals/corr/abs-2004-13845,DBLP:conf/nips/ZellersHRBFRC19}.
Yet, beside the prompt used to initiate the generation process, there are few options to have control on the generation process. Being able to add some constraints on the generated texts is useful for various situations. For example, it allows to create texts that follow a certain writing style, convey a certain emotion or polarity or to ensure that a generated summary contains correct information. 
More critically, it can be used to prevent the inherent toxicity of language models trained on the internet, or to not reproduce gender or race stereotypes. So far, most methods necessitate to fine-tune the LM, so that it specifically learns to model this constraint, i.e. the constraint is --hopefully-- incorporated in the LM.
This fine-tuning approach has several drawbacks. It implies to train multiple specific LMs (one per constraint), which is costly, when even possible given the size of current state-of-the-art LM, and results in several models. 

In this paper, we propose new approaches to add such additional constraints on the texts but at decoding time. We exploit a discriminator that is trained to determine if a text follows a given constraint or not; its output provides information to guide the generation toward texts that satisfy this expected constraint.
In order to make the most of the discriminator information, we propose an original method based on the Monte Carlo Tree Search (MCTS) algorithm \citep{DBLP:conf/cg/Coulom06}, namely Plug and Play Language - Monte Carlo Tree Search (PPL-MCTS).
We also propose simpler methods based on re-ranking to fulfil this goal.
Both approaches do not require to fine-tune the LM; adding a new constraint can thus simply be done by providing a discriminator verifying if a text complies with what is expected.
More precisely, our main contributions are the following ones:
\begin{enumerate}
    \item we propose to use MCTS as a decoding strategy to implement constrained generation and we show, on 3 datasets and 2 languages, that it yields state-of-the-art results while offering more flexibility;
    \item we also explore simpler generation methods based on re-ranking and show that this kind of approach, with low computational costs, can also be competitive if the diversity within propositions to re-rank is encouraged;
    \item we provide a fully functional code implementing a batched textual MCTS\footnote{\url{https://github.com/NohTow/PPL-MCTS}} working with the popular HuggingFace's Transformers library~\cite{DBLP:conf/emnlp/WolfDSCDMCRLFDS20}
\end{enumerate}


\section{Related work}
\label{sec:related-work}

The goal of constrained textual generation is to find the sequence of tokens $x_{1:T}$ which maximises $p(x_{1:T} \mid c)$, given a constraint $c$.
Few methods address the constrained textual generation.

\paragraph{Class-conditional language models.}
Class-conditional language models (CC-LMs), as the Conditional Transformer Language (CTRL) model \citep{DBLP:journals/corr/abs-1909-05858}, train or fine-tune the weights $\theta$ of a single neural model directly for controllable generation, by appending a control code in the beginning of a training sequence. The control code indicates the constraint to verify and is related to a class containing texts that satisfy the constraint. For the sake of simplicity, we will denote without distinction the class, the constraint verified by its texts and the associated control code by $c$.
Trained with different control codes, the model learns $p_\theta(x_{1:T} \mid c) = \prod_{t=1}^{T} p_\theta(x_{t} \mid x_{1:t-1}, c)$. The constraint can then be applied during generation by appending the corresponding control code to the prompt.
While this method gives some kind of control over the generation, the control codes need to be defined upfront and the LM still needs to be trained specifically for each set of control codes. This is an important limitation since the current trend in text generation is the use of large pre-trained models which can hardly be fine-tuned (for instance, the last version of GPT, GPT-3, cannot be fine-tuned without access to very large hardware resources).

\paragraph{Discriminator-based methods}
The general idea of discriminator-guided generation is to combine a disciminator $D$ with a generative LM. 
The discriminator explicitly models the constraint by calculating the probability $p_{D}(c \mid x_{1:T})$ of the sequence $x_{1:T}$ to satisfy the constraint $c$. This probability is directly related to $p(x_{1:T}\mid c)$ through Bayes' rule : $ p(x_{1:T}\mid c) \propto p_{D}(c \mid x_{1:T}) p_\theta(x_{1:T})$.
Discriminator-based methods alleviate the training cost problem, as discriminators are easier to train than a LM. Moreover, any additional constraint can be defined a posteriori without tuning the LM, only by training another discriminator.
The discriminators have been used in different ways to explore the search space. In the work of \citep{DBLP:conf/acl/ChoiBGHBF18, DBLP:conf/icml/ScialomDLPS20}, the space is first searched using beam search to generate a pool of proposals with a high likelihood $p_\theta(x_{1:T})$, and then the discriminator is used to re-rank them. 
However, in addition that beam search can miss sequences with high likelihood, it is biased towards the likelihood, while the best sequence might only have an average likelihood, but satisfies the constraint perfectly. 
 
Hence, it might be more suitable to take the discriminator probability into account during decoding rather than after generating a whole sequence. In this case, the discriminator is used at each generation step $t$ to get the probability $p_D(c \mid x_{1:t})$ for each token of the vocabulary $\mathcal{V}$, and merge it to the likelihood $p_{\theta}(x_{1:t})$ to choose which token to emit. 
In order to reduce the cost of using a discriminator on every possible continuation, 
GeDi~\citep{DBLP:journals/corr/abs-2009-06367} proposes to use CC-LMs as generative discriminators.
The method relies on the fact that the CC-LM computes $p_{\theta}\left(x_{t} \mid x_{1:t-1}, c\right)$ for all tokens of the vocabulary which can be used to get $p_{\theta}(c \mid x_{1: t})$ for all tokens using Bayes' equation.
This approach is thus at the intersection of tuning the LM and using a discriminator: it tunes a small LM (the CC-LM) to guide a bigger one.  


In Plug And Play Language Model (PPLM)~\citep{DBLP:conf/iclr/DathathriMLHFMY20}, the discriminator is used to shift the hidden states of the pre-trained transformer-based LM towards the desired class at every generation step.
PPLM can be used on any LM and with any discriminator.
However, PPLM needs to access the LM to modify its hidden states, while our approach only requires the output logits. As some LM can only be used through access to logits (e.g. GPT-3 API), this makes our approach more plug and play than PPLM.

A common drawback of all these approaches is their lack of a long-term vision of the generation. Indeed, the discriminator probabilities become necessarily more meaningful as the sequence grows and might only be trustable to guide the search when the sequence is (nearly) finished. When used in a myopic decoding strategy, classification errors will cause the generation process to deviate further and further. 
Trying to optimize a score defined in the long horizon by making short term decisions is very similar to common game setups such as chess, where the Monte Carlo Tree Search (MCTS) has proven to be really effective~\citep{Silver1140}, which motivated our approach.

\section{PPL-MCTS method}
\label{sec:methodology}

The approach that we propose is in line with methods using a discriminator to guide a large LM decoding, without the need to re-train it. Also, it can be applied to any LM with any discriminator, following the plug and play paradigm. Unlike previous approaches, it is able to have a long term vision on what is generated.
Being able to make a short-term decision (choice of the next token $x_t$ at time step $t$) that is promising in the long run is based on the exploration of the search space. We propose here to use the Monte Carlo Tree Search (MCTS) for an efficient exploration of this space. 

MCTS is very well suited for this problem for three reasons. First, it allows to get a local score (i.e, a score for the next token to emit) using finished sequences. Hence, this score is more meaningful than scores based only on the next step. Second, it allows to explicitly define the compromise between exploitation of promising sequences (with a high likelihood), and exploration of other potentially promising sequences (to not miss better sequences with a lower likelihood). The fact that regret, i.e the number of simulations done on a sub-optimal sequence, has a theoretical upper bound in MCTS~\citep{DBLP:journals/amai/Rosin11} is a nice guarantee that the computation time is not wasted and the search is efficient. Finally, it outputs a solution at each iteration and so can fit our computational budget by allowing to adjust the quality of the solution to calculation spent.

\begin{figure}
\centering
\includegraphics[width=0.8\columnwidth]{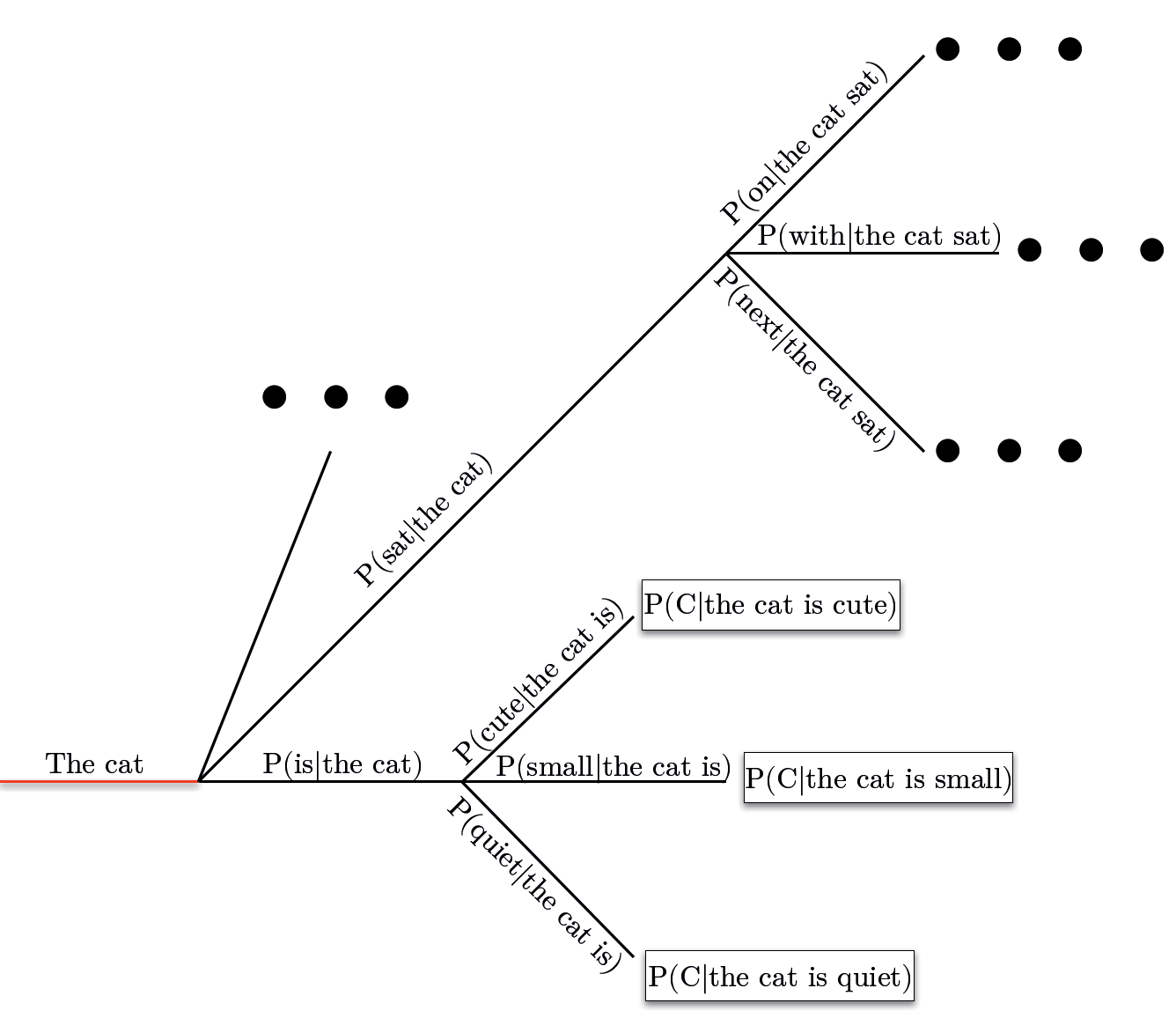} 
\caption{Illustration of the constrained generation process as a tree exploration from the prompt \texttt{The cat}. Classification probabilities are only represented on completed sequences.}
\label{fig:treeexplo}
\end{figure}

\paragraph{Text generation as tree exploration process.}
The search space of the text generation corresponds to a tree: its root is the prompt and the child of a node is its father's sequence with one of the $\vert \mathcal{V} \vert$ possible tokens appended.
In the case of constrained generation, the goal is thus to find the path, and therefore the sequence $x$, with the highest $p(x \mid c)$ possible without exploring the whole tree in width and depth. As mentioned previously, this probability can be computed as the product of the likelihood $p_{\theta}(x)$ and the probability given by a discriminator $p_D(c\mid x)$.
An illustration of such a tree can be found in Fig.~\ref{fig:treeexplo}, where the likelihood of $x$ is forged by multiplying corresponding conditional probabilities along the path, and the classification probability is calculated at the terminal node.

\paragraph{Applying MCTS to text generation.}
MCTS is a heuristic based iterative algorithm that uses randomness to solve deterministic problems that cannot be solved using traditional approaches, often because the search space is too large to be entirely explored. Each iteration consists in four consecutive steps.
In the particular context of applying MCTS to text generation, we made some adaptations: 

\begin{enumerate}
  \item \textbf{Selection} Recursively choose children from the root to a node that has not been expanded yet. To only explore viable sequences, the probability $p_{\theta}(x_i \mid x_{1:t-1})$ of a given token $x_i$ given by the LM is used during the selection phase. To this end, the children chosen are those maximizing the Polynomial Upper Confidence Trees (PUCT)~\citep{DBLP:journals/amai/Rosin11} as defined in~\citep{DBLP:journals/nature/SilverSSAHGHBLB17}: 
\begin{equation}
    \label{eqn:puct}
    PUCT(i) = \frac{s_{i}}{n_{i}} + c_{puct} \; p_{\theta}(x_i \mid x_{1:t-1})\frac{\sqrt{N_{i}}}{1+n_i}
\end{equation}
with $s_i$ the aggregated score of the node $i$, $n_{i}$ the number of simulations played after this node, $N_{i}$ the number of simulations played after its parent, and $c_{puct}$ a constant defining the compromise between exploration and exploitation.
In the task of constrained generation, we define the score of a sequence as its probability knowing the class $p(x \mid c)$. 

  \item \textbf{Expansion} If the selected node is not terminal, use the LM to expand it by creating its children. 
  \item \textbf{Simulation (roll-out)} Sample one of these children according to $p_{\theta}(x_i \mid x_{1:t-1})$, and go to a terminal node through a random walk or another pattern.
  \item \textbf{Backpropagation} Aggregate the final score obtained at the terminal node ($p(x \mid c)$) to each parent until root. There are different strategies to aggregate scores, as computing the average between the actual score and the one being backpropagated, or taking the maximum of the two. We take the aggregated score $s_i$ associated to the node $i$ as the averaged probability over all simulations played after this node.
\end{enumerate}

When the number of iterations has reached the allocated budget, the building of the tree stops. The token $x_i$ selected for the current decoding step can be selected as the most played node amongst the root’s children nodes, or the one with the highest aggregated score. We chose the most played one.

\begin{figure}[]
\includegraphics[width=\columnwidth]{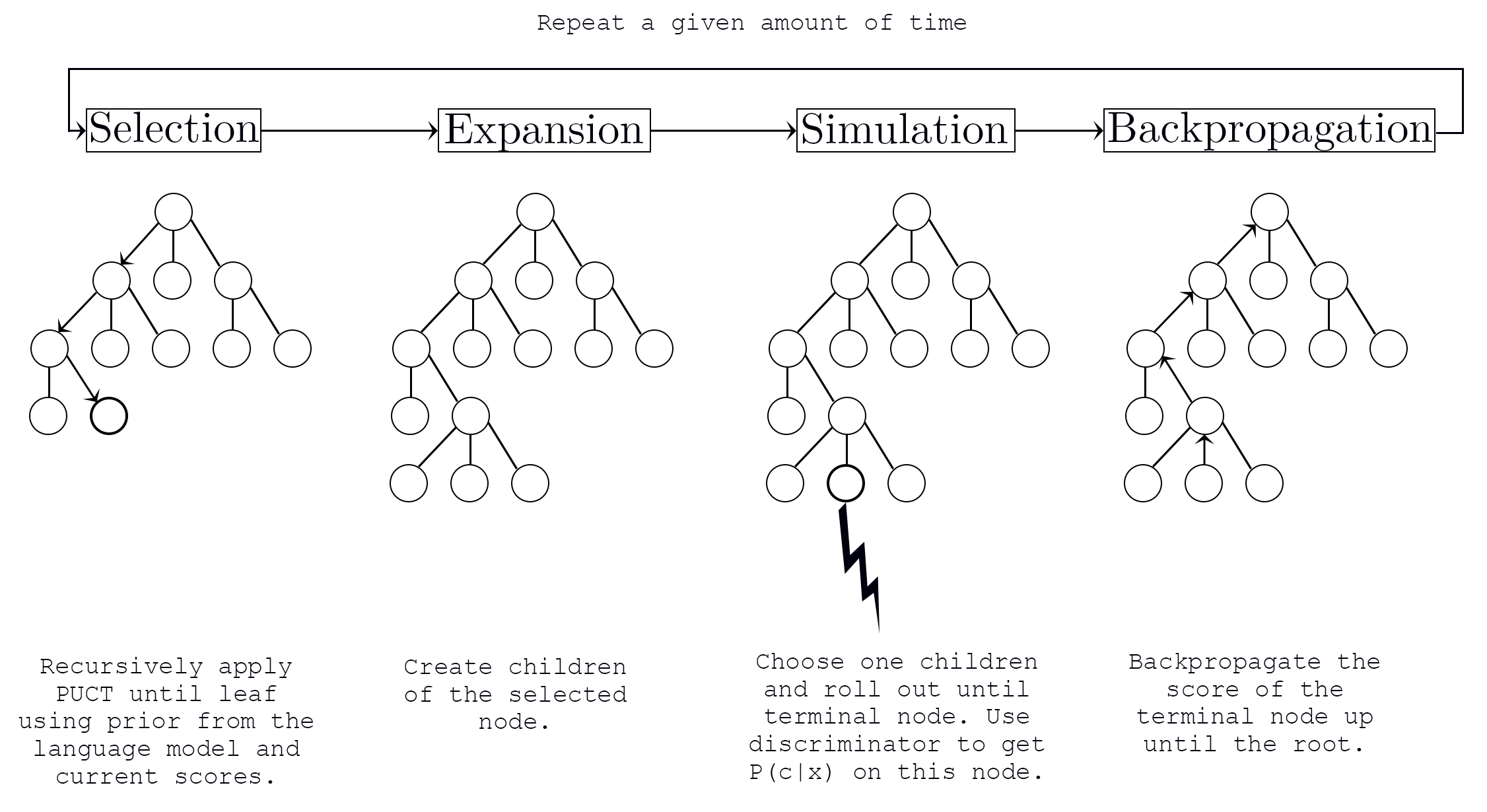} 
\caption{MCTS application to text generation.}
\label{fig:mcts}
\end{figure}

These adaptations of MCTS to constrained generation are summarized in Fig. \ref{fig:mcts}. Note that any language model can be used for defining the probability $p_{\theta}(x_i \mid x_{1:t-1})$ and any discriminator for scoring sequences, hence the name of our approach: Plug and Play Language - Monte Carlo Tree Search (PPL-MCTS).
MCTS has been very recently used for machine translation~\citep{DBLP:conf/emnlp/LeblondASPLASV21}, question generation and summarization~\citep{selfGAN}. The differences with these concurrent studies are discussed in Appendix~\ref{sec:concurrent_works}.


\paragraph{Model improvements.}

In order to allow a finer control on how the constraint is applied, we introduce a parameter $\alpha \in [0,1]$ to control the compromise between likelihood and constraint strength, modifying Bayes' equation: $p(x\mid c) \propto p_D(c \mid x)^{\alpha}p_{\theta}(x)^{1-\alpha}$. 
Note that PUCT (\ref{eqn:puct}) already considers the likelihood of the sequence, favoring the selection of nodes with high likelihoods. Hence, even sequences generated with $\alpha = 1$ are correctly written. Setting $\alpha < 1$ forces the algorithm to explore solutions even closer to the language model. In our experiments, we set $\alpha = 1$ to strengthen the constraint.

To avoid expensive roll-outs, one may also assign a value to unfinished sequences at the cost of a less precise evaluation that may be not as meaningful as when doing roll-outs. Indeed, the discriminator can be trained on sequences with variable numbers of tokens, allowing it to be used at each node without the need of simulations. In this setup, the MCTS is used as an efficient compromise between exploration and exploitation, losing part of its long view property 
but allowing to skew the exploration toward promising solutions.

Finally, during our first experiments, we observed that PPL-MCTS leads to repetitive patterns. This is very similar of what happens with greedy search, where a single sequence with a high likelihood is dominating the search. If such sequences also have a pretty high discriminator scores, they will be repeated often. CTRL~\citep{DBLP:journals/corr/abs-1909-05858} offers a simple yet very powerful method to avoid noisy repetitions. 
It applies a scalar factor $I(i)$ to the temperature parameter $\tau$ of a given token $x_i$ that penalizes this token if it is already in the input sequence. The probability of a given token becomes:
\begin{equation}
    \label{eqn:penalized_logit}
p_{\theta}^{'}(x_i \mid x_{1:t-1}) = \frac{\exp \left(z_i / (\tau \cdot I(i)))\right.}{\sum_{v} \exp \left(z_v / (\tau \cdot I(v) ))\right.}
\end{equation}
with the \textit{repetition penalty} $I(i)> 1$ if $x_i$ is already in the prompt and 1 otherwise, and $z$ the neural LM predicted logits over the vocabulary $\mathcal{V}$. Thus, probabilities of already emitted tokens are penalized, but if the language model gives a really high score to one token (hence, it is very confident that this \textit{should} be the token to emit), it may still be selected as the output token.

\section{Experiments}
\label{sec:experiments}

\subsection{Performance assessment}

The goal of constrained generation is to generate samples that 1) belong to a specific class while 2) keeping the language quality of the original LM, and 3) with enough diversity across samples. 
We chose three different metrics to evaluate each of these aspects: 1) accuracy, which is verified by an external "oracle" discriminator trained on a dataset disjoint from the one used to guide the generation; 2)  perplexity, which is computed using an "oracle" LM, i.e an unconstrained LM trained on different data than the one used to train the constrained generator; 3) Self-BLEU score \citep{DBLP:conf/sigir/ZhuLZGZWY18}, which is the BLEU score \citep{DBLP:conf/acl/PapineniRWZ02} of a sample using the other samples as references: a high Self-BLEU score means that there is a lot of overlap between generated samples, and thus that the diversity is low.
Such automatic metrics have known limitations~\citep{DBLP:conf/iclr/CacciaCFLPC20} but results of human evaluation on the CLS dataset, detailed in Section~\ref{sec:human_eval}, confirm that they provide a good overview of the performance. 


In practice, the studied dataset (see below) is split into two parts, each part being sub-divided in train/val/test sets. The first part serves to train models used for the generation (LM and discriminator), while the second is used to train oracles which serve to compute the automatic evaluation metrics. 
The test set of this second part will also be used to forge prompts for the generation. Further details on data splits are given in Appendix~\ref{sec:data_splits}.
Each metric is evaluated on a pool of 900 generated samples.

\subsection{Datasets}

Three different datasets are used in the experiments presented hereafter: \href{https://huggingface.co/datasets/amazon_polarity}{amazon\_polarity}~\citep{DBLP:conf/nips/ZhangZL15}, CLS (from the \href{https://huggingface.co/datasets/flue}{FLUE}~\citep{DBLP:conf/lrec/LeVFSCLACBS20} dataset) and \href{https://huggingface.co/datasets/emotion}{emotion}~\citep{DBLP:conf/emnlp/SaraviaLHWC18}.
The first two are Amazon reviews which have been labeled as positive or negative, so the intended task is to study the possibility of applying polarity to the generation. As CLS is in French, these two datasets will serve to ensure that the methods have the same behaviour for different languages.
Emotion is a collection of tweets classified under eight basic emotions: anger, anticipation, disgust, fear, joy, sadness, surprise and trust. This dataset is supposed to be more challenging since there are more classes and texts are smaller (only composed of one sentence), hence the model needs to precisely generate the target emotion with few tokens. 
It is worth noting that the 3 datasets have different sizes: 4,000,000 instances in total for amazon\_polarity, 20,000 for emotion and 6,000 for CLS. They are available at \url{https://huggingface.co/datasets/}.

We adapted prompts used to start the generation for each datasets depending on the data format. Amazon\_polarity comes with a "title" column which corresponds to the title the user gave to the review. This field is directly used as prompt. For the two other datasets, the prompts are the very first tokens of the text field. Because texts from emotion and CLS have different lengths, the size of prompts are also different: it is arbitrarily set to 6 tokens for CLS and 4 for emotion.

\subsection{Methods and baselines}


\paragraph{Baselines.}
Beside PPL-MCTS, we propose several baselines and simple techniques. 
Most studies on re-ranking create proposals using beam search and then re-rank them using the product of likelihood and discriminator probability, limiting the diversity in the proposals pool. 
We propose re-ranking with different variations, in the way sequences to re-rank are produced, and in the way the final sequence is chosen in an attempt to improve such approaches.
Three methods are tested to generate propositions: beam search~\citep{dept._2018} (with a beam size of 3), nucleus (top-p) sampling~\citep{DBLP:conf/iclr/HoltzmanBDFC20} (with p=0.9), as well as beam sampling (as described in~\citep{DBLP:conf/iclr/CacciaCFLPC20}). For the final choice, we also propose three different methods: \textit{argmax}, where the sequence that has the highest $p(x|c)$ is chosen; \textit{first true}, where propositions are sorted by descending likelihood and the first sequence that belongs to the correct class according to the guiding discriminator is chosen; and \textit{sampling}, where the distribution of $p(x|c)$ for the propositions is normalized and the chosen sequence is sampled following this distribution. Similarly to PPL-MCTS, the likelihood part of $p(x|c)$ is omitted (i.e, $\alpha=1$) since sequences in the pool of propositions already have an high likelihood. 

It should be noted that in our setting, a generated sequence corresponds to a document (e.g. a whole review). 
This choice makes sense for our datasets, but re-ranking at a smaller level (after each sentence, after x tokens...) would also be possible and might produce different results.

\paragraph{Methods from the literature}

We compare our results with methods from the literature.
In particular, we test CC-LMs trained on the target task, similarly as CTRL. 
We tested this method using greedy search as well as sampling for decoding. 
We also propose an implementation of CC-LM trained with the classification loss initially proposed for the GeDi method \citep{DBLP:journals/corr/abs-2009-06367}.
These CC-LMs are further used to implement the state-of-the-art GeDi model. In the experiments reported below, we report results for GeDi models trained with and without the classification loss.
Finally, we report results of PPLM. For a fair comparison, the same discriminator and LM are used for our PPL-MCTS approach, the re-ranking approaches (baselines), and PPLM.

\subsection{Experimental setting}

For each method, a number of tokens equal to the average length of sequences of the dataset are generated: 98 tokens for amazon\_polarity, 23 for emotion and 137 for CLS. Fixing the number of generated tokens ensures fair comparisons between the tested methods. Indeed, even though perplexity and Self-BLEU metrics are normalized by the length of the sequence, these metrics can tend to penalize a model producing longer sequences: such model has more risk to deviate and repeat itself, which would results in higher values compared to a model producing shorter sequences.
An example of generation from amazon\_polarity is given in Fig.~\ref{fig:ex_generation}.

\begin{figure}

    \centering
    \fbox{%
    \parbox{0.99\columnwidth}{%
        \scriptsize  
\textbf{<|startoftext|> The Revenge of making a good Halloween film. [SEP]}?????? I think this movie is a waste of time. It's not scary, it's just plain stupid. The only good thing about this film is the soundtrack.<|endoftext|>\newline
\textbf{<|startoftext|> The Revenge of making a good Halloween film. [SEP]} ive seen this movie a few times and i love it. the acting is great, the story line is good, and the special effects are awesome. if you like horror movies then go see this one.<|endoftext|>
}}
\caption{Example of two constrained generations using PPL-MCTS, one on the negative class, one on the positive class, using the same prompt (in bold) from amazon\_polarity.}
\label{fig:ex_generation}

\end{figure}


\begin{figure}
\includegraphics[width=\columnwidth]{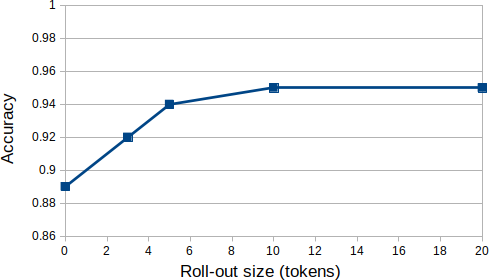}
    \caption{Accuracy according to the roll-out size; CLS dataset}
    \label{fig:rollout}


\end{figure}

To run all of these methods, three different models are needed: one discriminator, a "vanilla" LM used as generator, and the CC-LM used in the CTRL and GeDi approaches. For the discriminator used to guide the generation, we rely on \href{https://huggingface.co/bert-base-cased}{BERT-base-cased}~\citep{DBLP:conf/naacl/DevlinCLT19} for the English datasets and \href{https://huggingface.co/flaubert/flaubert_large_cased}{FlauBERT-large-cased}~\citep{DBLP:conf/lrec/LeVFSCLACBS20} for CLS.  As vanilla LM, we use GPT-2 small models, relying on OpenAI's \href{https://huggingface.co/gpt2}{pre-trained model} for the English datasets and on \href{https://huggingface.co/antoiloui/belgpt2}{belgpt2} for the French one. The implementation and models used for BERT, FlauBERT, GPT-2 and belgpt2 are all found on \url{https://huggingface.co/models}. 
Given the particular format of data on our experimental datasets, the vanilla LM is trained on raw training sequences in order to produce texts corresponding to the task (for instance, reviews). 
The CC-LM is simply a fine-tuned version of the vanilla LM with the control code appended. 

We tested three values for the temperature parameter for each proposed method (1.0, 1.1 and 1.2).
For PPL-MCTS, we also studied the impact of $c_{puct}$ by testing values 1.0, 3.0, 5.0 and 8.0 along with the different temperature values mentioned. We only report the results for parameters yielding the best accuracy score in the main paper but every results can be found in Appendix~\ref{sec:comp_results}.
The repetition penalty has been set to 1.2 as defined in CTRL. 
The number of MCTS iteration per token is set to 50, as well as the number of propositions for re-ranking, except for beam sampling where it is set to 10 because of memory limitations. 
Given the cost of roll-out for long sequences, we apply roll-out only on the emotion dataset to be able to run extensive experiments. Without roll-out, MCTS loses a part of its long view property but still allows to skew the exploration toward promising solutions.
A study of the impact of the roll-out is detailed in a next sub-section.
Parameters used for literature models are those provided by the authors. Experiments were conducted on a Quadro RTX 6000 with 80 GB of RAM.

\subsection{Results}

Results on the emotion, CLS and amazon\_polarity datasets are reported in Table~\ref{tab:all_datasets}. The statistical significance against GeDi and PPLM is measured with a t-test with significance level (p-value) of 1\%.
Results show that PPL-MCTS is competitive against task-specifically trained LMs on the constraint application aspect (high accuracy), while keeping a fair amount of diversity (low Self-BLEU) and staying close to the original distribution (low oracle perplexity). 
On all three datasets and metrics, it constantly yields top results; the only other method which is high-performing for all metrics and constant across the datasets is GeDi trained with the classification loss.


Another remarkable result is for the Sampling - Argmax method that selects among a pool of propositions generated using sampling, the one with the highest probability to be from the correct class. Thanks to the sampling used for generating propositions, its Self-BLEU is among the lowest of all reported values. Despite the simplicity and low computational cost of this approach, its accuracy is among the best on every dataset. These very good results should however be put into perspective of the very high perplexity of its generated texts. 
This indicates that the generated samples may be very different than those generated by a standard LM on this dataset.
Hence, exploring accuracy/perplexity trade-offs achievable with different values of $\alpha$ is interesting, which is proposed in Appendix~\ref{sec:constraint_strength}.

\begin{table*}[htb]
\resizebox{\textwidth}{!}{
\begin{tabular}{llcccccccc}
                                         & \multicolumn{3}{c}{amazon\_polarity}                                                                                   & \multicolumn{3}{c}{emotion}                                                                                 & \multicolumn{3}{c}{CLS}                                                                 \\
\multicolumn{1}{l|}{Generation}          & \multicolumn{1}{c}{Accuracy ↑}        & 5 - Self-BLEU  ↓                      & \multicolumn{1}{c|}{Oracle}                         & Accuracy ↑                 & 5 - Self-BLEU  ↓                       & \multicolumn{1}{c|}{Oracle}                         & Accuracy ↑                 & 5 - Self-BLEU  ↓                        & Oracle                         \\
\multicolumn{1}{l|}{method}              & \multicolumn{1}{c}{}           &                         & \multicolumn{1}{c|}{perplexity ↓}                        &                     &                         & \multicolumn{1}{c|}{perplexity ↓}                        &                     &                          & perplexity ↓                        \\ \hline

\multicolumn{1}{l|}{\textbf{Tuned LM}}   &                            &                            & \multicolumn{1}{c|}{} &  &                 & \multicolumn{1}{c|}{}          &           &  &                            \\

\multicolumn{1}{l|}{CC-LM - Classloss}   & 0.82                            & 0.79                           & \multicolumn{1}{c|}{\textbf{2.56}$^{\ast,\dagger}$} & \textbf{0.89}$^\ast$ & 0.65$^\dagger$                 & \multicolumn{1}{c|}{3.72$^{\ast,\dagger}$}          & 0.89$^\ast$          & \textbf{0.04}$^{\ast, \dagger}$ & 50.6                           \\
\multicolumn{1}{l|}{CC-LM}               & 0.91                            & 0.71                           & \multicolumn{1}{c|}{3.21$^\dagger$}                 & 0.52                 & \textbf{0.13}$^{\ast,\dagger}$ & \multicolumn{1}{c|}{11.1}                           & 0.66                 & 0.06$^{\ast, \dagger}$          & 31.5                           \\ \hline
\multicolumn{1}{l|}{GeDi - Classloss}    & 0.96$^\ast$                     & 0.6$^\ast$                     & \multicolumn{1}{c|}{5.16}                           & 0.88$^\ast$          & 0.68                           & \multicolumn{1}{c|}{5.57$^\ast$}                    & 0.94$^\ast$          & 0.4                             & 7.99$^\ast$                    \\
\multicolumn{1}{l|}{GeDi}                & 0.96$^\ast$                     & 0.6$^\ast$                     & \multicolumn{1}{c|}{5.16}                           & 0.54                 & 0.52$^\dagger$                 & \multicolumn{1}{c|}{4.09$^{\ast,\dagger}$}          & 0.83$^\ast$          & 0.31$^\dagger$                  & 11.9                           \\ \hline
\multicolumn{1}{l|}{\textbf{Untuned LM}}   &                            &                            & \multicolumn{1}{c|}{} &  &                 & \multicolumn{1}{c|}{}          &           &  &                            \\
\multicolumn{1}{l|}{PPLM}                & 0.89                            & 0.66                           & \multicolumn{1}{c|}{2.84$^\dagger$}                 & 0.67                 & 0.19$^\dagger$                 & \multicolumn{1}{c|}{7.31}                           & 0.79                 & 0.23$^\dagger$                  & 8.3                            \\ \hline
\multicolumn{1}{l|}{Beam - Argmax}     & 0.88                            & 0.85                           & \multicolumn{1}{c|}{3.14$^\dagger$}                 & 0.72$^\ast$          & 0.49$^\dagger$                 & \multicolumn{1}{c|}{3.7$^{\ast,\dagger}$}           & 0.64                 & 0.82                            & 3.31$^{\ast,\dagger}$          \\
\multicolumn{1}{l|}{Beam - Sampling}   & 0.86                            & 0.84                           & \multicolumn{1}{c|}{3.27$^\dagger$}                 & 0.7                  & 0.46$^\dagger$                 & \multicolumn{1}{c|}{3.69$^{\ast,\dagger}$}          & 0.6                  & 0.82                            & 3.37$^{\ast,\dagger}$          \\
\multicolumn{1}{l|}{Beam - First true} & 0.85                            & 0.83                           & \multicolumn{1}{c|}{3.27$^\dagger$}                 & 0.65                 & 0.38$^\dagger$                 & \multicolumn{1}{c|}{\textbf{3.68}$^{\ast,\dagger}$} & 0.62                 & 0.82                            & \textbf{3.26}$^{\ast,\dagger}$ \\ \hline
\multicolumn{1}{l|}{Beam sampling - Argmax}     & 0.97$^\ast$                     & 0.73                           & \multicolumn{1}{c|}{3.82$^\dagger$}                 & 0.67                 & 0.48$^\dagger$                 & \multicolumn{1}{c|}{3.88$^{\ast,\dagger}$}          & 0.88$^\ast$          & 0.67                            & 3.91$^{\ast,\dagger}$          \\
\multicolumn{1}{l|}{Beam sampling - Sampling}   & 0.92                            & 0.76                           & \multicolumn{1}{c|}{3.68$^\dagger$}                 & 0.66                 & 0.48$^\dagger$                 & \multicolumn{1}{c|}{3.88$^{\ast,\dagger}$}          & 0.76                 & 0.63                            & 4.07$^{\ast,\dagger}$          \\
\multicolumn{1}{l|}{Beam sampling - First true} & 0.9                             & 0.73                           & \multicolumn{1}{c|}{3.84$^\dagger$}                 & 0.66                 & 0.49$^\dagger$                 & \multicolumn{1}{c|}{3.85$^{\ast,\dagger}$}          & 0.85$^\ast$          & 0.71                            & 3.8$^{\ast,\dagger}$           \\ \hline
\multicolumn{1}{l|}{Sampling - Argmax}     & \textbf{0.99}$^{\ast, \dagger}$ & 0.17$^{\ast,\dagger}$          & \multicolumn{1}{c|}{16.5}                           & 0.87$^\ast$          & \textbf{0.13}$^{\ast,\dagger}$ & \multicolumn{1}{c|}{11.7}                           & 0.92$^\ast$          & 0.12$^{\ast, \dagger}$          & 14.3                           \\
\multicolumn{1}{l|}{Sampling - First true} & 0.89                            & \textbf{0.07}$^{\ast,\dagger}$ & \multicolumn{1}{c|}{85.9}                           & 0.82$^\ast$          & \textbf{0.13}$^{\ast,\dagger}$ & \multicolumn{1}{c|}{10.4}                           & 0.87$^\ast$          & 0.14$^{\ast, \dagger}$          & 13                             \\
\multicolumn{1}{l|}{Sampling - Sampling}   & 0.88                            & 0.17$^{\ast,\dagger}$          & \multicolumn{1}{c|}{16.3}                           & 0.81$^\ast$          & \textbf{0.13}$^{\ast,\dagger}$ & \multicolumn{1}{c|}{10.4}                           & 0.81                 & 0.06$^{\ast, \dagger}$          & 31.8                           \\ \hline
\multicolumn{1}{l|}{PPL-MCTS}            & 0.97$^\ast$                     & 0.63$^\ast$                    & \multicolumn{1}{c|}{5.69}                           & 0.84$^\ast$          & 0.37$^\dagger$                 & \multicolumn{1}{c|}{4.82$^{\ast,\dagger}$}          & 0.89$^\ast$          & 0.54                            & 4.98$^{\ast,\dagger}$         \\
\multicolumn{1}{l|}{PPL-MCTS - 10 tokens roll-out}            &                     &                     & \multicolumn{1}{c|}{}                           &           &                  & \multicolumn{1}{c|}{}          & \textbf{0.95}$^\ast$          & 0.57                            & 5.07$^{\ast,\dagger}$   
\end{tabular}
}
\caption{Performance of constrained generation methods; from left to right: amazon\_polarity, emotion, CLS datasets. $\dagger$ (resp. $\ast$) indicates statistically significant improvement against GeDi-classloss (resp. PPLM).}  
\label{tab:all_datasets}
\end{table*}

\subsection{Human evaluation}
\label{sec:human_eval}
Since automatic metrics can be biased and may not faithfully represent the human judgement, we conduct a human evaluation to compare with the results obtained through oracles and confirm the results and the relevance of automatic metrics.
Because of the annotation cost, we limit the tested methods to the two state-of-the-art methods (PPLM and GeDi), PPL-MCTS and the promising Sampling - Argmax. This allows to test if PPL-MCTS is indeed as efficient as GeDi and if both are better than original PPLM. 
Also, this should confirm that the high perplexity of the Sampling - Argmax method is due to generated texts being very different from the ones generated by other methods. 
The evaluation has been performed on the CLS dataset by three volunteering colleagues, French native speakers.
They labeled the same pool of reviews in order to measure the inter-annotator agreement. 

The pool consists of 50 reviews (25 positive and 25 negative ones) randomly sampled for each method, which results in 200 reviews in total. Annotators were asked to go through this (randomly shuffled) pool and to give two scores for each review: 
\begin{enumerate}
    \item \textbf{Polarity}. Rate from 1 to 5 how well the text corresponds to the desired label (positive or negative).
    The text is rated 5 if it corresponds entirely to the expected label, down to 1 if it corresponds entirely to the opposite label.
    This score corresponds to the accuracy from the automatic metrics.
    \item \textbf{Readability}. Rate from 1 to 5 how well the text is written. 5 corresponds to a text without any mistake and which is perfectly understandable. 
    The more mistakes or incoherence, the lower the score. 
    This score corresponds to the perplexity from the automatic metrics.
\end{enumerate}
The diversity within the pool of generated texts is complicated to evaluate and the Self-BLEU is fairly accurate as a diversity metric, so this property is not studied in the human evaluation.

We report scores averaged over the 3 annotators as well as the standard deviation in Table~\ref{tab:human_eval}. A t-test against PPLM (GeDi being best on both scores) is applied to test statistical significance (with p-value=0.01). 
One can notice that the agreement between annotators is high and that the results are in line with conclusions from automatic metrics. GeDi, when trained with the classification loss, yields similar results as PPL-MCTS, in terms of constraint satisfaction and quality of writing. PPLM, on the other hand, generates samples of lower quality and has more difficulty for applying the constraint. Finally, given its readability score, Sampling - Argmax seems to generate samples with a low quality. Its polarity score, while being higher than PPLM, is lower than expected: given the accuracy reported by the oracle, it should be close to GeDi and PPL-MCTS. It is most likely due to the fact that evaluating the polarity of a badly written text is hard for an human, often resulting in review being scored as neutral.

\begin{table}[]
\resizebox{\columnwidth}{!}{
\centering
\begin{tabular}{@{}lll@{}}
\toprule
Generation method & Polarity        & Readability     \\ \midrule
GeDi - Classloss  & $4,46 \pm 0,08^\ast$ & $4,19 \pm 0,28^\ast$ \\
PPL-MCTS          & $4,43 \pm 0,12^\ast$ & $4,05 \pm 0,23^\ast$ \\
PPLM              & $3,74 \pm 0,08$ & $3,12 \pm 0,19$ \\
Sampling - Argmax & $4,00 \pm 0,11$ & $2,83 \pm 0,33$ \\ \bottomrule
\end{tabular}}
\caption{Results of the human evaluation on the CLS dataset (averaged over 3 annotators). $\ast$ indicates statistically significant ($p \leq 1\%$) improvement against PPLM.}
\label{tab:human_eval}
\end{table}

\subsection{Effect of the roll-out}
Rolling out is costly for very long sequences, and the question of its usefulness necessarily arises. 
We study how rolling out for only a fixed number of tokens (instead of until the end of the sequence) influences the performance of PPL-MCTS. 
For this experiment, we use the CLS dataset and set the roll-out to 0 (original result), 3, 5, 10 and 20 tokens.
As one can note in Fig.~\ref{fig:rollout}, only 5 tokens allows PPL-MCTS to be on par with GeDi on this dataset. The roll-out size quickly improves accuracy, which then reaches a plateau. It suggests that having an horizon is really helpful but only up to a certain point. Beside, Self-BLEU and oracle perplexity values stay stable, varying respectively from 0.54 to 0.57, and from 4.98 to 5.18 as the roll-out size increases from 0 to 20.
The roll-out size can thus be set accordingly to the compute budget, further defining the trade-off between cost and quality. 

 
%
%

\section{Conclusion}

In this paper, we show that it is possible to control generation with the help of a discriminator that implements some expected constraints on the text during decoding. 
This flexible approach is very useful when using very large language models, such as GPT-3, whose fine-tuning computational costs are prohibitive. In contrast, training a discriminator is easier and cheaper.
Our proposed methods, that mix the discriminator constraint and the generation, yield performance that is equivalent to the best approaches based on LM tuning at lower training cost.
On the other hand, such approaches have an additional cost during inference because of the cost of the discriminator being applied to candidate generations. A study on this additional cost depending on the type of discriminator used can be found in~\cite{DBLP:journals/corr/abs-2204-11586}. PPL-MCTS offers a solution for cases where training is too costly for the downstream application or the language model is not directly accessible.
Seeing text generation as a tree exploration process, an existing approach such as GeDi indeed lowers the cost of width exploration but the depth exploration is still an issue. Using GeDi for constrained generation is thus very similar to a standard maximum likelihood search which still lacks of an optimal search method.
On the other hand, Monte Carlo Tree Search provides an efficient way to explore the tree by determining the best local choice in the long run, lowering the cost of depth exploration. Thus, these two methods solve different facets of constrained generation, and the combination of the two is a promising perspective. 
Moreover, MCTS allows to precisely define the best compromise between cost and quality through the number of iterations and the roll-out size, while ensuring the efficiency of the search theoretically. 
For reproducibility purposes, our implementation is made available at \url{https://github.com/NohTow/PPL-MCTS}. \vspace{2mm} \newline 
\indent Several research avenues are opened by this work. For methods yielding high perplexity, it would be interesting to explore how to set the $\alpha$ parameter in order to reach the best compromise between accuracy and perplexity. Similarly, the size (number of tokens considered) of the roll-out in MCTS offers some ways to control the cost/performance compromise. An adaptive roll-out size, for example rolling-out until the score of the discriminator is above or below a threshold as in~\citep{Cotarelo2021}, would seem particularly suited for texts.
It should also be noted that fine-tuning a model and controlling the generation with a discriminator can be used jointly. For instance, one can use PPL-MCTS on a tuned LM, which will most likely result in even better performances because sequences considered during the search will have an overall higher quality for the considered task.
Finally, not only can PPL-MCTS be applied to any property that a discriminator can identify, but it can also work using other scoring methods (human evaluation, regular expressions, heuristic based evaluation, ...) as long as the score reflects compliance with the expected property.



\section{Ethics/Broader impact}

The ethical risks of large LMs are well known \citep{Bender2021}. Especially when they are trained on large quantities of non curated data, it has be shown that they tend to reproduce or amplifies biases on gender, race, etc. and more generally may produce inappropriate content \citep{Gehman2020}.
As for every automatic generation method, using our approaches may result in the production of unwanted, misleading or inappropriate content.
Yet, it is noteworthy that the constrained generation as we propose is one way to control, a posteriori of the LM training, that the generated texts respect some criteria. It can be used for any application given that a discriminator is able to check the constraint accurately. The ethical interests are thus important, such as adding constraint about race diversity, gender equality, non toxicity, factual faithfulness, etc. as far as these properties can be detected by a (trained or hand-crafted) discriminator.
But of course, the same technique could be used for malicious purposes, such as constraining generation so it produces offensive texts, targeted fake news, etc. In such cases of misuse, discriminators similar to those used for constraining the generation could easily spot such texts since the constraint will, by design, be noticeable and easily grasped by a discriminator. 

Even though training language models on curated data in the first place is possible, totally curated dataset is hard to obtain, and new biases may be highlighted. Indeed, defining a priori what is every possible bias in every cultural context for every possible application, and curating the training data accordingly is hardly feasible.
Hence, constant updates of language models will be necessary to make them as fair as possible. Given the cost of large language models training, updating them often is really harmful for the environment. Discriminator guided constrained generation offers a way to filter text generation using up-to-date standards in terms of fairness by only updating the discriminator, which is faster and require way less resources.


%

\bibliographystyle{acl_natbib}
\bibliography{biblio-generation}  

\clearpage
\appendix
\section{Appendix}


In this technical appendix, we provide additional information about our methods, some settings and the experiments. Further experimental results, as well as examples, are given and discussed. Finally, a discussion on concurrent studies is provided.

\subsection{Data splits}
\label{sec:data_splits}
We adapted the way we split the dataset into two parts and train/test/validation sets depending on the original dataset splits. Amazon\_polarity is composed of a training set of 3\,600\,000 examples and a test set of 400\,000. We split both into two parts and kept 20\% of each training set for validation.
Emotion already comes with train, test and validation set, hence we just split each into two parts. Finally, CLS is composed of a train set and a test set of 6000 examples. We split the training set in two and split the test set twice so we got two validation and test sets.
Thus, for each dataset, we end up with two training sets, two validation sets and two test sets.

The first train and validation sets are used to train and control the training of models used for the generation: the guiding classifier, the "vanilla" LM and the CC-LM. The test set serves to control their performance.

The second ones are used to train the LM oracle and the classifier used to measure the accuracy. The test set allows to verify that these models are trustworthy for accurate evaluation.
Once all the models are trained, the constrained generation is evaluated on 900 samples generated from prompts never seen by models during training.




\subsection{Complementary results}
\label{sec:comp_results}

We tested three temperature values for each proposed method: 1.0, 1.1 and 1.2. As the temperature increases, the output distribution of the language model becomes more and more uniform. This means that high temperatures should result in high perplexities because the sampling deviates further from the original distribution.

For PPL-MCTS, we also studied the impact of $c_{puct}$ by testing values 1.0, 3.0, 5.0 and 8.0 along with the different temperature values mentioned. $c_{puct}$ defines the compromise between exploiting nodes that already have great scores and exploring less played but promising ones. 
A high $c_{puct}$ encourages exploration.
We remind that the repetition penalty $I$ in Eqn.~\ref{eqn:penalized_logit} has been set to 1.2 as defined in CTRL. 

In Section 'Results', for each method and dataset, we reported only the results obtained with the set of parameter values yielding the best accurracy.
Hereafter, we report results with every tested set of parameters in Tables~\ref{tab:emotion_full},~\ref{tab:CLS_full} and~\ref{tab:amazon_full} for respectively the emotion, CLS and amazon\_polarity datasets. 

Unsurprisingly, the perplexity of methods which sample on the LM logits explodes when $\tau$ increases, without a noticeable gain in accuracy. Since the diversity is already high for low $\tau$ values, it seems to be better to keep the temperature low with these approaches. Note that the couple $c_{puct} = 3, \tau = 1.0$ for PPL-MCTS always leads to the best result. Using $c_{puct} = 8$ seems to yield slightly worse results, especially with a low temperature. However, the different parameters do not greatly affect the results of PPL-MCTS.


\subsection{Examples of generation}

We provide an example of generation for amazon\_polarity and emotion datasets using PPL-MCTS, PPLM, GeDi and Sampling - Argmax methods, respectively in Figures~\ref{fig:gens_amazons_appendix} and~\ref{fig:gens_emotion_appendix}. Texts generated using Sampling - Argmax are rather different as suggested by the reported high perplexity results. Note that emotion texts are only one sentence while those of amazon\_polarity are complete reviews. This difference motivated the choice of these datasets. Also, we preferred amazon\_polarity over IMDb used in the GeDi and PPLM papers because of its bigger size, suitable format and because a French equivalent is available (CLS), which allows us to test another language with a similar dataset.

\begin{figure}[]
    \centering
    \fbox{%
    \parbox{\columnwidth}{%
        \scriptsize  
    {\small \textbf{PPL-MCTS}} \newline   
\textbf{<|startoftext|> realistic account [SEP]} ive read this book and i think it is a very good account of the life of a young man. i would recommend to anyone who likes to know about the lives of people in america, or any other country that has been around for centuries.<|endoftext|>\newline\newline
   {\small \textbf{PPLM}}\newline
\textbf{<|startoftext|> realistic account [SEP]} I enjoyed this book. It was realistic and I enjoyed the way the author described the people and places. I would recommend this book to anyone who is interested in the history of the United States.<|endoftext|>\newline\newline
   {\small \textbf{GeDi}}\newline
\textbf{<|startoftext|> realistic account [SEP]} This is a real book. It's not just about the war, but also about the people who fought it. The author has a very realistic view of what happened in Vietnam and how it affected our country. I would recommend this book to anyone interested in learning more about Vietnam or any other war that took place during that time period<|endoftext|>\newline\newline
   {\small \textbf{Sampling - Argmax}}\newline
\textbf{<|startoftext|> realistic account [SEP]} ive seen many of these. they mostly in the 20's, 30's and 40's and the luscious ones i know - men like this - there's an old saying that farts, u makin' u sell it..this movie has a lot of realism to it too! and i was totally impressed on how good the kids and the predator was! will it be hard for them to make more like this? i think it will! i read that war is going to be much<|endoftext|>\newline

}}
\caption{Examples of constrained generation using PPL-MCTS, PPLM, GeDi and Sampling - Argmax methods (from top to bottom) on the positive class of amazon\_polarity, using the same prompt (in bold).}
\label{fig:gens_amazons_appendix}
\end{figure}


\begin{figure}[]
    \centering
    \fbox{%
    \parbox{\columnwidth}{%
        \scriptsize 
          {\small \textbf{PPL-MCTS}} \newline 
\textbf{<|startoftext|> i feel that working} with a group of people who are so passionate about the same thing is really important<|endoftext|>
\newline \newline
   {\small \textbf{PPLM}}\newline
\textbf{<|startoftext|> i feel that working} hard and caring for someone i don t care for is a lot less selfish than i would be feeling for someone i<|endoftext|>\newline\newline
{\small \textbf{GeDi}}\newline
\textbf{<|startoftext|> i feel that working} with the ladies of the family is a wonderful thing and i am very fond of the way they look and feel<|endoftext|>\newline\newline
{\small \textbf{Sampling - Argmax}}\newline
\textbf{<|startoftext|> i feel that working} at imgur for so many years is ill be devoted to it<|endoftext|>\newline

}}
\caption{Examples of constrained generation using PPL-MCTS, PPLM, GeDi and Sampling - Argmax methods (from top to bottom) on the 'love' class form 'emotion', using the same prompt (in bold).}
\label{fig:gens_emotion_appendix}
\end{figure}

\subsection{Constraint strength through $\alpha$}
\label{sec:constraint_strength}
As described in the model improvements section,  
a parameter $\alpha$ can be defined to control the relative importance of the discriminator score and the language model likelihood. Thus, this parameter allows to control the constraint application strength as it helps to define a trade-off between staying close the original LM and satisfying the constraint. Note that in all of our experiments reported earlier, this parameter has been set to 1, focusing on the constraint application since the proposed methods already inherently provide legible texts.

Here, as a proof of concept, we test a range of values for $\alpha$, using the Sampling - Argmax method on the amazon\_polarity dataset with the automatic metrics. We chose this method and dataset since it yields the best accuracy, but also exhibits a very high perplexity. In this case, it seems interesting to trade a bit of accuracy for better written texts.

\begin{figure*}[]
    \centering
 
%
%
\includegraphics[width=\textwidth]{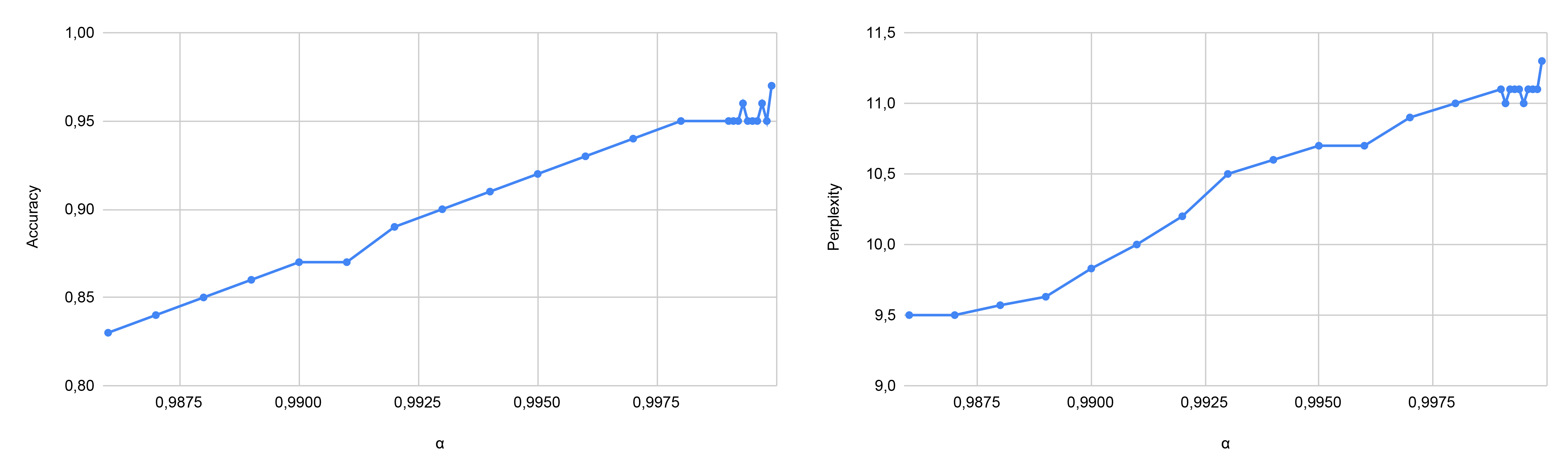}
    \caption{Accuracy (left) and perplexity (right) of the Sampling - Argmax method according to the $\alpha$ parameter; amazon\_polarity dataset}
    \label{fig:alpha}
\end{figure*}

Results are roughly constant when $\alpha$ is lower than 0.98, so it has an impact only for values between 0.98 and 1. This is due to the fact that, for a long enough sequence, $p_{\theta}(x)$ is often relatively small compared to $p_D(c \mid x)$. This difference of scale annihilates the influence of $\alpha$. This [0.98-1] interval thus corresponds to values of $\alpha$ that rescale $p_D(c \mid x)^{\alpha}$ and $p_{\theta}(x)^{1-\alpha}$ on a same order of magnitude. 
As shown in Figure~\ref{fig:alpha}, within this regime, we can observe a linear dependency between $\alpha$ and the accuracy as well as the perplexity. This illustrate that a trade-off can be obtained by tuning this parameter, allowing to define the strength of the constraint application which also defines how far the generation can be from the original LM distribution. 

\begin{table}[h]
\resizebox{\columnwidth}{!}{
\begin{tabular}{@{}lrrr@{}}
\toprule
Generation method                                & \multicolumn{1}{l}{Accuracy ↑} & \multicolumn{1}{l}{5 - Self-Bleu  ↓} & \multicolumn{1}{l}{Oracle perplexity ↓} \\ \midrule
Beam sampling - Argmax $\tau = 1.0$              & 0,61                           & 0,41                                 & 3,7                                     \\
Beam sampling - Argmax $\tau = 1.1$              & 0,65                           & 0,48                                 & 3,72                                    \\
\textit{Beam sampling - Argmax $\tau = 1.2$}     & \textit{0,67}                  & \textit{0,48}                        & \textit{3,88}                           \\ \midrule
Beam sampling - First true $\tau = 1.0$          & 0,58                           & 0,4                                  & 3,68                                    \\
Beam sampling - First true $\tau = 1.1$          & 0,64                           & 0,48                                 & 3,69                                    \\
\textit{Beam sampling - First true $\tau = 1.2$} & \textit{0,66}                  & \textit{0,49}                        & \textit{3,85}                           \\ \midrule
Beam sampling - Sampling $\tau = 1.0$            & 0,59                           & 0,41                                 & 3,69                                    \\
Beam sampling - Sampling $\tau = 1.1$            & 0,64                           & 0,49                                 & 3,69                                    \\
\textit{Beam sampling - Sampling $\tau = 1.2$}   & \textit{0,66}                  & \textit{0,48}                        & \textit{3,88}                           \\ \midrule
CC-LM - Greedy Search                            & 0,51                           & 0,1                                  & 17                                      \\
\textit{CC-LM - Sampling $\tau = 1.0$}           & \textit{0,52}                  & \textit{0,13}                        & \textit{11,1}                           \\
CC-LM - Sampling $\tau = 1.1$                    & 0,51                           & 0,1                                  & 15,8                                    \\
CC-LM - Sampling $\tau = 1.2$                    & 0,47                           & 0,08                                 & 31,4                                    \\ \midrule
\textit{CC-LM - Classloss - Greedy Search}       & \textit{0,89}                  & \textit{0,65}                        & \textit{3,72}                           \\
CC-LM - Classloss - Sampling $\tau = 1.0$        & 0,83                           & 0,11                                 & 19,6                                    \\
CC-LM - Classloss - Sampling $\tau = 1.1$        & 0,79                           & 0,07                                 & 33,2                                    \\
CC-LM - Classloss - Sampling $\tau = 1.2$        & 0,79                           & 0,05                                 & 64,8                                    \\ \midrule
\textit{Sampling - Argmax $\tau = 1.0$}          & \textit{0,87}                  & \textit{0,13}                        & \textit{11,7}                           \\
Sampling - Argmax $\tau = 1.1$                   & 0,86                           & 0,1                                  & 19,6                                    \\
Sampling - Argmax $\tau = 1.2$                   & 0,86                           & 0,07                                 & 47,5                                    \\ \midrule
\textit{Sampling - First true $\tau = 1.0$}      & \textit{0,82}                  & \textit{0,13}                        & \textit{10,4}                           \\
Sampling - First true $\tau = 1.1$               & 0,81                           & 0,11                                 & 16,2                                    \\
Sampling - First true $\tau = 1.2$               & 0,77                           & 0,09                                 & 33,2                                    \\ \midrule
\textit{Sampling - Sampling $\tau = 1.0$}        & \textit{0,81}                  & \textit{0,13}                        & \textit{10,4}                           \\
Sampling - Sampling $\tau = 1.1$                 & 0,8                            & 0,11                                 & 15                                      \\
Sampling - Sampling $\tau = 1.2$                 & 0,79                           & 0,08                                 & 25,7                                    \\ \midrule
PPL-MCTS - $c_{puct}= 1.0, \tau = 1.0$            & 0,83                           & 0,37                                 & 4,81                                    \\
PPL-MCTS - $c_{puct}= 1.0, \tau = 1.1$            & 0,8                            & 0,36                                 & 4,9                                     \\
PPL-MCTS - $c_{puct}= 1.0, \tau = 1.2$            & 0,82                           & 0,33                                 & 4,97                                    \\
\textit{PPL-MCTS - $c_{puct}= 3.0, \tau = 1.0$}   & \textit{0,84}                  & \textit{0,37}                        & \textit{4,82}                           \\
PPL-MCTS - $c_{puct}= 3.0, \tau = 1.1$            & 0,82                           & 0,35                                 & 4,85                                    \\
PPL-MCTS - $c_{puct}= 3.0, \tau = 1.2$            & 0,84                           & 0,35                                 & 4,9                                     \\
PPL-MCTS - $c_{puct}= 5.0, \tau = 1.0$            & 0,84                           & 0,38                                 & 4,74                                    \\
PPL-MCTS - $c_{puct}= 5.0, \tau = 1.1$            & 0,84                           & 0,34                                 & 4,79                                    \\
PPL-MCTS - $c_{puct}= 5.0, \tau = 1.2$            & 0,84                           & 0,33                                 & 4,88                                    \\
PPL-MCTS - $c_{puct}= 8.0, \tau = 1.0$            & 0,81                           & 0,38                                 & 4,71                                    \\
PPL-MCTS - $c_{puct}= 8.0, \tau = 1.1$            & 0,81                           & 0,37                                 & 4,72                                    \\
PPL-MCTS - $c_{puct}= 8.0, \tau = 1.2$            & 0,82                           & 0,35                                 & 4,79                    \\               
\end{tabular}}
\caption{Results for every tested set of parameters on the proposed methods; emotion dataset. Results reported in the body of the paper are in italic.}
\label{tab:emotion_full}
\end{table}

\begin{table}[h]
\resizebox{\columnwidth}{!}{
\begin{tabular}{@{}lrrr@{}}
\toprule
Generation method                                  & \multicolumn{1}{l}{Accuracy ↑} & \multicolumn{1}{l}{5 - Self-Bleu  ↓} & \multicolumn{1}{l}{Oracle perplexity ↓} \\ \midrule
Beam sampling - Argmax $\tau = 1.0$                & 0,87                           & 0,71                                 & 3,85                                    \\
\textit{Beam sampling - Argmax $\tau = 1.1$}       & \textit{0,88}                  & \textit{0,67}                        & \textit{3,91}                           \\
Beam sampling - Argmax $\tau = 1.2$                & 0,88                           & 0,63                                 & 4,12                                    \\ \midrule
\textit{Beam sampling - First true $\tau = 1.0$}   & \textit{0,85}                  & \textit{0,71}                        & \textit{3,8}                            \\
Beam sampling - First true $\tau = 1.1$            & 0,84                           & 0,68                                 & 3,87                                    \\
Beam sampling - First true $\tau = 1.2$            & 0,85                           & 0,63                                 & 4,07                                    \\ \midrule
Beam sampling - Sampling $\tau = 1.0$              & 0,74                           & 0,71                                 & 3,82                                    \\
Beam sampling - Sampling $\tau = 1.1$              & 0,72                           & 0,68                                 & 3,89                                    \\
\textit{Beam sampling - Sampling $\tau = 1.2$}     & \textit{0,76}                  & \textit{0,63}                        & \textit{4,07}                           \\ \midrule
CC-LM - Greedy Search                              & 0,59                           & 0,57                                 & 2,51                                    \\
CC-LM - Sampling $\tau = 1.0$                      & 0,62                           & 0,15                                 & 12,3                                    \\
CC-LM - Sampling $\tau = 1.1$                      & 0,63                           & 0,09                                 & 18,7                                    \\
\textit{CC-LM - Sampling $\tau = 1.2$}             & \textit{0,66}                  & \textit{0,06}                        & \textit{31,5}                           \\ \midrule
CC-LM - Classloss - Greedy Search                  & 0,8                            & 0,59                                 & 2,77                                    \\
CC-LM - Classloss - Sampling $\tau = 1.0$          & 0,85                           & 0,13                                 & 17                                      \\
CC-LM - Classloss - Sampling $\tau = 1.1$          & 0,87                           & 0,07                                 & 28                                      \\
\textit{CC-LM - Classloss - Sampling $\tau = 1.2$} & \textit{0,89}                  & \textit{0,04}                        & \textit{50,6}                           \\ \midrule
\textit{Sampling - Argmax $\tau = 1.0$}            & \textit{0,92}                  & \textit{0,12}                        & \textit{14,3}                           \\
Sampling - Argmax $\tau = 1.1$                     & 0,92                           & 0,08                                 & 20,7                                    \\
Sampling - Argmax $\tau = 1.2$                     & 0,92                           & 0,05                                 & 33,6                                    \\ \midrule
\textit{Sampling - First true $\tau = 1.0$}        & \textit{0,87}                  & \textit{0,14}                        & \textit{13}                             \\
Sampling - First true $\tau = 1.1$                 & 0,86                           & 0,1                                  & 19,1                                    \\
Sampling - First true $\tau = 1.2$                 & 0,86                           & 0,06                                 & 33,1                                    \\ \midrule
Sampling - Sampling $\tau = 1.0$                   & 0,77                           & 0,14                                 & 12,9                                    \\
Sampling - Sampling $\tau = 1.1$                   & 0,78                           & 0,09                                 & 18,8                                    \\
\textit{Sampling - Sampling $\tau = 1.2$}          & \textit{0,81}                  & \textit{0,06}                        & \textit{31,8}                           \\ \midrule
PPL-MCTS - $c_{puct}= 1.0, \tau = 1.0$              & 0,88                           & 0,54                                 & 4,98                                    \\
PPL-MCTS - $c_{puct}= 1.0, \tau = 1.1$              & 0,87                           & 0,53                                 & 5                                       \\
PPL-MCTS - $c_{puct}= 1.0, \tau = 1.2$              & 0,87                           & 0,53                                 & 5,02                                    \\
\textit{PPL-MCTS - $c_{puct}= 3.0, \tau = 1.0$}     & \textit{0,89}                  & \textit{0,54}                        & \textit{4,98}                           \\
PPL-MCTS - $c_{puct}= 3.0, \tau = 1.1$              & 0,89                           & 0,54                                 & 4,81                                    \\
PPL-MCTS - $c_{puct}= 3.0, \tau = 1.2$              & 0,89                           & 0,54                                 & 4,86                                    \\
PPL-MCTS - $c_{puct}= 5.0, \tau = 1.0$              & 0,88                           & 0,55                                 & 4,9                                     \\
PPL-MCTS - $c_{puct}= 5.0, \tau = 1.1$              & 0,89                           & 0,54                                 & 4,97                                    \\
PPL-MCTS - $c_{puct}= 5.0, \tau = 1.2$              & 0,89                           & 0,54                                 & 4,91                                    \\
PPL-MCTS - $c_{puct}= 8.0, \tau = 1.0$              & 0,83                           & 0,54                                 & 4,98                                    \\
PPL-MCTS - $c_{puct}= 8.0, \tau = 1.1$              & 0,86                           & 0,54                                 & 4,95                                    \\
PPL-MCTS - $c_{puct}= 8.0, \tau = 1.2$              & 0,88                           & 0,55                                 & 4,94             \\                      
\end{tabular}}
\caption{Results for every tested set of parameters on the proposed methods; CLS dataset. Results reported in the body of the paper are in italic.} 
\label{tab:CLS_full}
\end{table}

\begin{table}[h]
\resizebox{\columnwidth}{!}{
\begin{tabular}{@{}lrrr@{}}
\toprule
Generation method                                & \multicolumn{1}{l}{Accuracy ↑} & \multicolumn{1}{l}{5 - Self-Bleu  ↓} & \multicolumn{1}{l}{Oracle perplexity ↓} \\ \midrule
Beam sampling - Argmax $\tau = 1.0$              & 0,94                           & 0,79                                 & 3,55                                    \\
Beam sampling - Argmax $\tau = 1.1$              & 0,96                           & 0,77                                 & 3,65                                    \\
\textit{Beam sampling - Argmax $\tau = 1.2$}     & \textit{0,97}                  & \textit{0,73}                        & \textit{3,82}                           \\ \midrule
Beam sampling - First true $\tau = 1.0$          & 0,86                           & 0,77                                 & 3,73                                    \\
Beam sampling - First true $\tau = 1.1$          & 0,89                           & 0,77                                 & 3,68                                    \\
\textit{Beam sampling - First true $\tau = 1.2$} & \textit{0,9}                   & \textit{0,73}                        & \textit{3,84}                           \\ \midrule
Beam sampling - Sampling $\tau = 1.0$            & 0,87                           & 0,77                                 & 3,7                                     \\
\textit{Beam sampling - Sampling $\tau = 1.1$}   & \textit{0,92}                  & \textit{0,76}                        & \textit{3,68}                           \\
Beam sampling - Sampling $\tau = 1.2$            & 0,89                           & 0,73                                 & 3,83                                    \\ \midrule
\textit{CC-LM - Greedy Search}                   & \textit{0,91}                  & \textit{0,71}                        & \textit{3,21}                           \\
CC-LM - Sampling $\tau = 1.0$                    & 0,87                           & 0,17                                 & 15,7                                    \\
CC-LM - Sampling $\tau = 1.1$                    & 0,86                           & 0,1                                  & 32,2                                    \\
CC-LM - Sampling $\tau = 1.2$                    & 0,8                            & 0,08                                 & 80,2                                    \\ \midrule
\textit{CC-LM - Classloss - Greedy Search}       & \textit{0,82}                  & \textit{0,79}                        & \textit{2,56}                           \\
CC-LM - Classloss - Sampling $\tau = 1.0$        & 0,81                           & 0,16                                 & 18,4                                    \\
CC-LM - Classloss - Sampling $\tau = 1.1$        & 0,79                           & 0,1                                  & 37,1                                    \\
CC-LM - Classloss - Sampling $\tau = 1.2$        & 0,74                           & 0,07                                 & 95,4                                    \\ \midrule
\textit{Sampling - Argmax $\tau = 1.0$}          & \textit{0,99}                  & \textit{0,17}                        & \textit{16,5}                           \\
Sampling - Argmax $\tau = 1.1$                   & 0,99                           & 0,11                                 & 31,8                                    \\
Sampling - Argmax $\tau = 1.2$                   & 0,99                           & 0,07                                 & 84,50                                   \\ \midrule
Sampling - First true $\tau = 1.0$               & 0,88                           & 0,16                                 & 16,4                                    \\
Sampling - First true $\tau = 1.1$               & 0,87                           & 0,1                                  & 31,5                                    \\
\textit{Sampling - First true $\tau = 1.2$}      & \textit{0,89}                  & \textit{0,07}                        & \textit{85,9}                           \\ \midrule
\textit{Sampling - Sampling $\tau = 1.0$}        & \textit{0,88}                  & \textit{0,17}                        & \textit{16,3}                           \\
Sampling - Sampling $\tau = 1.1$                 & 0,87                           & 0,1                                  & 30,8                                    \\
Sampling - Sampling $\tau = 1.2$                 & 0,88                           & 0,07                                 & 81                                      \\ \midrule
PPL-MCTS - $c_{puct}= 1.0, \tau = 1.0$            & 0,96                           & 0,62                                 & 5,61                                    \\
PPL-MCTS - $c_{puct}= 1.0, \tau = 1.1$            & 0,96                           & 0,63                                 & 5,65                                    \\
PPL-MCTS - $c_{puct}= 1.0, \tau = 1.2$            & 0,96                           & 0,62                                 & 5,66                                    \\
\textit{PPL-MCTS - $c_{puct}= 3.0, \tau = 1.0$}   & \textit{0,97}                  & \textit{0,63}                        & \textit{5,69}                           \\
PPL-MCTS - $c_{puct}= 3.0, \tau = 1.1$            & 0,97                           & 0,62                                 & 5,77                                    \\
PPL-MCTS - $c_{puct}= 3.0, \tau = 1.2$            & 0,96                           & 0,62                                 & 5,72                                    \\
PPL-MCTS - $c_{puct}= 5.0, \tau = 1.0$            & 0,95                           & 0,63                                 & 5,6                                     \\
PPL-MCTS - $c_{puct}= 5.0, \tau = 1.1$            & 0,96                           & 0,63                                 & 5,66                                    \\
PPL-MCTS - $c_{puct}= 5.0, \tau = 1.2$            & 0,96                           & 0,63                                 & 5,63                                    \\
PPL-MCTS - $c_{puct}= 8.0, \tau = 1.0$            & 0,93                           & 0,64                                 & 5,57                                    \\
PPL-MCTS - $c_{puct}= 8.0, \tau = 1.1$            & 0,93                           & 0,64                                 & 5,57                                    \\
PPL-MCTS - $c_{puct}= 8.0, \tau = 1.2$            & 0,95                           & 0,63                                 & 5,57           \\                        
\end{tabular}}
\caption{Results for every tested set of parameters on the proposed methods; amazon\_polarity dataset. Results reported in the body of the paper are in italic.} 
\label{tab:amazon_full}
\end{table}

\subsection{Concurrent work}
\label{sec:concurrent_works}

During the time of writing, two preprints using MCTS for NLP tasks have been released~\cite{DBLP:conf/emnlp/LeblondASPLASV21, selfGAN}. While we emphasize that these are concurrent studies, PPL-MCTS has some major differences.
Indeed, these studies focus on improving the overall quality of generated texts rather than following a given constraint. While "being well written" can be seen as a constraint, PPL-MCTS rather explores how a constraint that is not present in the original language model (i.e. not a goal in the original training of the LM) can be added at generation time. 
\citet{selfGAN} train a discriminator to distinguish generated and real samples because their goal is ultimately to train the language model in a Generative Adversarial setup to create a better LM. This iterative training, in addition to not being possible in our task, is not wanted since we aim to be plug and play. 
Our goal is indeed to apply an additional constraint to an untouched original language model. Yet, even if goals are different and applying MCTS for constrained generation is not trivial, the "MLE Coop-MCTS" is close to PPL-MCTS. However, focusing on MCTS as a decoding only strategy allowed an in-depth study that provided interesting results, in particular the effect of the roll-out size (the roll-out is totally omitted in their paper) and the $\alpha$ parameter.

On the other hand, \citet{DBLP:conf/emnlp/LeblondASPLASV21} also focus on MCTS as a decoding strategy but for the very specific case of machine translation. MCTS is used to optimize metrics for machine translation, which are known to not necessarily correlate with human judgement~\citep{DBLP:conf/emnlp/NovikovaDCR17}. 
Again, the goal is different since these metrics are used as a proxy of the sample quality. 
In contrast, our work shows that MCTS can be used to optimize a given property, but instead of optimizing the quality of samples, we optimize for a given constraint while retaining the original quality of writing. The fact that MCTS also works in such cases was non trivial since adding such constraints to the generation could lead to deteriorate texts.

Beside MCTS, we also proposed and explored simpler methods based on re-ranking for our task and showed that diversity allows to satisfy the constraint, often at the price of a lower quality, emphasizing the compromise between exploration and exploitation made by the MCTS.

Finally, these concurrent studies provide evidences that MCTS is promising for many different usage in NLP. We hope that the large amount of experiments, parameter analysis and the availability of our open-sourced code working out-of-the-box will help foster future research in this direction.

\end{document}